\documentclass[letterpaper, 10 pt, journal, twoside]{IEEEtran} 
\IEEEoverridecommandlockouts    

\let\labelindent\relax

\usepackage[table,usenames,dvipsnames]{xcolor}      
\usepackage{extarrows}                              
\usepackage{enumitem}
\usepackage{wrapfig}

\usepackage{amsmath,amssymb,amsfonts,amsthm,dsfont} 
\usepackage{algorithm,algorithmicx,listings}        
\usepackage[noend]{algpseudocode}			              
\makeatletter
\def\BState{\State\hskip-\ALG@thistlm}
\makeatother

\usepackage{graphicx}
\usepackage{tabularx}
\usepackage{subcaption}
\usepackage[font={small}]{caption}   
\usepackage[breaklinks=true, colorlinks, bookmarks=true, citecolor=Black, urlcolor=Violet,linkcolor=Black]{hyperref}

\usepackage[colorinlistoftodos,prependcaption,textwidth=1.6cm,textsize=normal]{todonotes}
\usepackage[normalem]{ulem}
\usepackage[utf8]{inputenc}
\usepackage[english]{babel}
\usepackage{hyperref}
\hypersetup{
    colorlinks=true,
    linkcolor=blue,
    filecolor=magenta,      
    urlcolor=cyan,
}

\newcommand{\prl}[1]{\mathopen{}\left(#1\right)\mathclose{}}

\DeclareMathAlphabet\mathbfcal{OMS}{cmsy}{b}{n}

\newtheorem*{assumption*}{Assumption}

\newtheorem*{problem*}{Problem}

\usepackage{tikz}
\usepackage{lipsum}

\newcommand\copyrighttext{%
  \footnotesize \centering \copyright 2019 IEEE. Personal use of this material is permitted. Permission from IEEE must be obtained for all other uses, \\ in any current or future media, including reprinting/republishing this material for advertising or promotional purposes,  creating new \\ collective works, for resale or redistribution to servers or lists, or reuse of any copyrighted component of this work in other works.}
\newcommand\copyrightnotice{%
\begin{tikzpicture}[remember picture,overlay]
\node[anchor=south,yshift=10pt] at (current page.south) {{\parbox{\dimexpr \textwidth-\fboxsep-\fboxrule\relax}{\copyrighttext}}};
\end{tikzpicture}%
}

\begin{document}

\title{Monocular Camera Based Fruit Counting and Mapping with Semantic Data Association}

\author{Xu Liu, Steven W. Chen, Chenhao Liu, Shreyas S. Shivakumar, Jnaneshwar Das, \\ Camillo J. Taylor, James Underwood, Vijay Kumar
\thanks{Manuscript received September 10, 2018; Revised December 9, 2018; Accepted January 23, 2019.}
\thanks{This paper was recommended for publication by Editor Youngjin Choi upon evaluation of the Associate Editor and the Reviewers' comments. This work was supported by ARL DCIST CRA grant W911NF-17-2-0181, ARO grant W911NF-13-1-0350, USDA grant 2015-67021-23857, NSF grant IIS-1138847, Qualcomm Research, C-BRIC (a Semiconductor Research Corporation Joint University Microelectronics Program cosponsored by DARPA), and the Australian Centre for Field Robotics (ACFR) at The University of Sydney, with thanks to Simpsons Farms.} 
\thanks{X. Liu,~~S. W. Chen,~~C. Liu,~~S. S. Shivakumar,~~C. J. Taylor,~~and V. Kumar are with GRASP Lab, University of Pennsylvania, Philadelphia, PA 19104, USA, {\tt\small\{liuxu, chenste, liuch13, sshreyas, cjtaylor, kumar\}@seas.upenn.edu}.} 
\thanks{J. Das is with the School of Earth and Space Exploration, Arizona State University, Tempe, AZ 85281, USA, {\tt\small jnaneshwar.das@asu.edu}.}
\thanks{J. Underwood is with the Australian Centre for Field Robotics, The
University of Sydney, 2006, Australia, {\tt\small j.underwood
@acfr.usyd.edu.au}.
}
\thanks{Digital Object Identifier (DOI): see top of this page.}
}
\markboth{IEEE Robotics and Automation Letters. Preprint Version. Accepted JANUARY, 2019}
{Liu \MakeLowercase{\textit{et al.}}: Monocular Camera Based Fruit Counting and Mapping}  

\maketitle

\copyrightnotice


\begin{abstract}
We present a cheap, lightweight, and fast fruit counting pipeline. Our pipeline relies only on a monocular camera, and achieves counting performance comparable to a state-of-the-art fruit counting system that utilizes an expensive sensor suite including a monocular camera, LiDAR and GPS/INS on a mango dataset. Our pipeline begins with a fruit and tree trunk detection component that uses state-of-the-art convolutional neural networks (CNNs). It then tracks fruits and tree trunks across images, with a Kalman Filter fusing measurements from the CNN detectors and an optical flow estimator. Finally, fruit count and map are estimated by an efficient fruit-as-feature semantic structure from motion (SfM) algorithm which converts 2D tracks of fruits and trunks into 3D landmarks, and uses these landmarks to identify double counting scenarios. There are many benefits of developing such a low cost and lightweight fruit counting system, including applicability to agriculture in developing countries, where monetary constraints or unstructured environments necessitate cheaper hardware solutions.\\


\end{abstract}

\section{Introduction}
\label{sec:intro}

\IEEEPARstart{A}{ccurately} estimating fruit count is important for growers to optimize yield and make decisions for harvest scheduling, labor allocation, and storage. 
Robotic fruit counting systems typically utilize a variety of sensors such as stereo cameras, depth sensors, LiDAR, and global positioning inertial navigation systems (GPS/INS). These systems have demonstrated great success in counting a variety of fruits including mangoes, oranges, and apples~\cite{das2015devices},~\cite{bargoti2017deep},~\cite{stein2016image}. However, while the use of a variety of high-end sensors results in good counting accuracy, they come at high monetary, weight, and size costs. For example, a sensor suite equipped with cameras, LiDAR, and a computer can add up to about $\$25,000$, and weigh upwards of a few kilograms~\cite{das2015devices}. 

These high monetary, weight, and size costs directly limit the applicability of these systems. Calibration of sensors poses additional challenges, when multiple sensing modalities such as cameras and LiDAR are used. A key motivation of this work is to develop a fruit counting system for cashew growers in Mozambique. The lack of infrastructure and technical knowledge, and tight cost constraints make it infeasible to use a complex sensor suite in these agriculture environments. The growth of smartphone technology have made high-quality monocular cameras readily available and accessible. These factors motivate the development of a high-performance fruit counting pipeline that uses only a monocular camera, and can potentially run on smartphones. By doing so, we would like to shift the burden of performance from sophisticated hardware, to sophisticated algorithms on cheap and ubiquitous commodity hardware.

\begin{figure}[t!]
\centering
\includegraphics[width=2.3in]{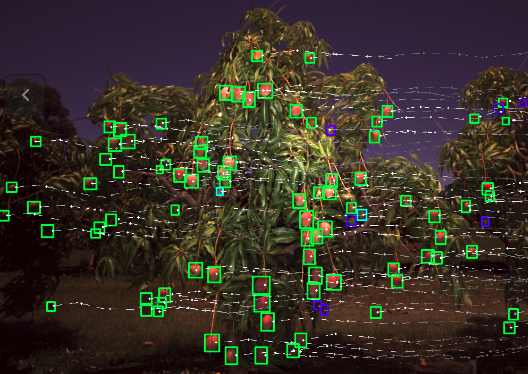}
\caption{{\textit{Detection and tracking of mangoes across images}}. The Faster R-CNN is used to detect fruits. Green boxes and blue boxes respectively represent stably tracked fruits and newly detected fruits. Every dotted white line represents the trajectory of a fruit. Each fruit will finally be associated with a 3D landmark in order to obtain the total count and the map of fruits.}
\label{fig:front_page}
\vspace*{-0.3in}
\end{figure}

The main \textbf{contributions} of our work are: (1) a monocular fruit counting system, which uses semantic SfM and improves upon our previous work~\cite{liu2018robust} by directly reconstructing fruit landmarks as opposed to geometric features; and (2) a thorough comparison on a mango dataset to a fruit counting system that uses additional sensors including a monocular camera, LiDAR and GPS/INS~\cite{stein2016image}, demonstrating that our counting system can achieve comparable performance using only the monocular camera data. Fig.~\ref{fig:front_page} depicts the detection and tracking performance of our algorithm. A video of our algorithm can be found at: \url{https://label.ag/ral19.mp4}.


\section{Related Work}


Fruit detection, segmentation and counting in a single image has seen a revolution from hand-crafted computer vision techniques to data-driven techniques. Traditional hand-engineered techniques for this task is usually based on a combination of shape detection and color segmentation. Dorj et al. develop a watershed segmentation based method to detect citrus in HSV space~\cite{dorj2017yield}. Ramos et al. use contour analysis on superpixel over-segmentation result to fit ellipses for counting coffee fruits on branches~\cite{ramos2017automatic}. Roy et al. develop a two-step apple counting method which first uses RGB-based oversegmentation for fruit area proposal, then estimates fruit count by fitting a clustering model with different center numbers~\cite{roy2018vision}. Such hand-crafted features usually have difficulty generalizing to different datasets where illumination or occlusion level may be different.

%
%
%
%

Consequently, data-driven methods have become the state of the art, primarily as a result of advances in deep learning methods. Bargoti et.al use Faster Region based Convolutional Neural Networks (Faster R-CNN) in detection of mangoes, almonds and apples, while also providing a standardized dataset for evaluation of counting algorithms~\cite{bargoti2017deep,bargoti2017image}. Chen et al. use the Fully Convolutional Network (FCN) to segment the image into candidate regions and CNN to count the fruit within each region~\cite{chen2017}. Rahnemoonfar and Sheppard train an Inception style architecture to directly count the number of tomatoes in an image, demonstrating that in some scenarios these deep networks can even be trained using synthetic data~\cite{rahnemoonfar2017deep}. Barth et al. also generate synthetic data to train deep neural networks to segment pepper images~\cite{barth2018data}. 

Our work differs from these previous works by expanding the counting problem from a single image to image sequences. Also, we limit ourselves to using only a monocular camera, which presents additional challenges over previous works including our chosen benchmark algorithm in~\cite{stein2016image}, since depth and pose information are not directly available from a LiDAR or GPS/INS sensors. 

%
%

Compared with counting fruits in a single image, in a structure orchard, counting fruits in two image sequences recorded from two opposite sides of every tree row is a more complete and accurate yield estimation approach, since most fruits are only visible from a certain viewpoint. However, this approach is more challenging because it introduces three major double counting problems: (1) double counting the same fruit detected in consecutive images; (2) double counting the same fruit visible from both sides of the tree; and (3) double counting fruits that are initially tracked, then lost, and then detected and tracked again in a later image, which is also called double tracking.

To overcome those challenges, existing approaches use some combinations of SfM, Hungarian algorithm, optical flow, and Kalman filters to provide the corresponding assignments of fruits across subsequent images. Wang et al. use stereo cameras to count red and green apples by taking images at night in order to control the illumination to exploit specular reflection features~\cite{wang2013automated}. Das et al. use a Support Vector Machine (SVM) to detect fruits, and use optical flow to associate the fruits in between subsequent images~\cite{das2015devices}. Halstead et al. use a 2D tracking algorithm to track and refine Faster R-CNN detections of sweet peppers for counting and crop quantity evaluation in an indoor environment~\cite{halstead2018fruit}. Roy et al. develop a four-step 3D reconstruction method which first roughly aligns 3D point cloud two-side view of fruit tree row, then generates semantic representation with deep learning-based trunk segmentations and further refines two-view alignment with this data. At back-end it uses the 3-D point cloud and pre-detected fruits from~\cite{roy2018vision} to give both visual count and tree height and size estimation for harvest count estimation~\cite{dong2018semantic}. 


Our previous monocular camera based fruit counting approach first maintains fruit tracks in the 2D image plane across frames. Separately, to reject outlier fruits, a computationally expensive SfM reconstruction is performed using SIFT features~\cite{liu2018robust}. The major improvement from this work lies in: (1) a fruit-as-feature semantic SfM is proposed, which significantly reduces the computation and outputs a meaningful map of fruit landmarks; (2) these fruit landmarks are then used to identify double tracked fruits, and eliminate frame-to-frame tracking noise caused by illumination shifts; (3) a new double counting problem introduced by collecting two separate image sequences from two opposite sides of the tree row is solved, which makes this pipeline applicable to a wider range of fruit counting tasks; (4) a more consistent and robust Kalman Filter scheme for frame-to-frame tracking is designed; and (5) a thorough comparison against both actual field counts and the counts estimated by a benchmark algorithm using much more expensive sensors is conducted.

\section{Fruit Detection with Deep Learning}

\begin{figure*}[t!]
\centering
\includegraphics[width=7.0in]{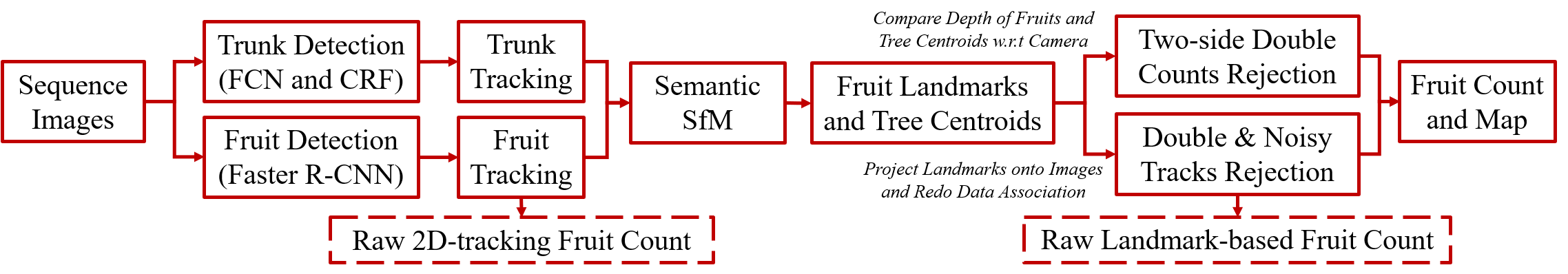}
\vspace*{-0.1in}
\caption{\textit{Proposed pipeline}. To count fruits on a specific row of trees, our pipeline takes in two image sequences recorded from two opposite sides of the tree row. Firstly, fruits and tree trunks are detected using CNNs. These detected fruits and trunks are then tracked across images, with a Kalman Filter fusing measurements from the CNN detector and the optical flow estimator. The semantic SfM process directly takes in these fruit and trunk tracks. It then uses tracked fruits' centers as feature matches across images, and estimates fruit and trunk landmark positions and camera poses. The two-side double counted fruits are identified by comparing the depth wrt the camera center of every fruit landmark and that of its corresponding tree centroid (approximated by the trunk landmark). Next, fruit data association is re-conducted by matching fruit landmarks' re-projections and fruit detections in every image, which automatically identifies double tracked fruits and eliminates noise in 2D tracking. Finally, we further refine the fruit data association and estimate the total count and map of fruits.}
\label{fig:pipeline}
\vspace*{-0.25in}
\end{figure*}

Our fruit detection component takes in sequence images, and outputs bounding boxes of fruits in each image as shown in Fig.\ref{fig:front_page}. The fruit detection is based on Faster R-CNN~\cite{ren2015faster}. The Faster R-CNN framework consists of two modules. The first module is a region proposal network which detects regions of interest. The second module is a classification module, which classifies individual regions and regresses the bounding box for every fruit simultaneously. Finally, probability threshold is applied and non-maximum suppression is conducted to remove duplicate detections.


We follow the methodology by Bargoti et al. for obtaining ground truth annotations for network training~\cite{bargoti2017deep}. These annotations are obtained by randomly sampling 1500 cropped images of size 500 $\times$ 500 from all 15,000 images of the orchard, each with original size of $3296 \times 2472$ pixels (8.14 megapixels). For ground truth, each fruit is labeled as a rectangular bounding box, giving both size and location information. Only fruits on trees in the first row are labeled. Labels of fruits on the ground truth trees are excluded from the training set. A Python-based annotation toolbox is publicly available at~\cite{clapperpython}. We refer the reader to~\cite{bargoti2017deep,stein2016image} for more details in implementation and performance of the fruit detection component.

\section{Fruit Counting with Landmark Representation}

The fruit detections in each image frame are used to construct a fruit count for each tree. The challenge in this step is associating detections with each other across all the image frames in the entire dataset, or in other words, identifying double counts. These associated detections then represent a single fruit.


We consider three kinds of situations which lead to double counts. The first results from observing the same fruit across consecutive images, which we address by tracking fruits in the 2D image plane. The second is double tracking caused by the re-observation of a previously tracked fruit. The third is double counting of fruits visible from two opposite views of the tree (i.e. robot facing east and facing west).


Our fruit counting and mapping pipeline thus mainly consists of five parts. The first part performs 2D tracking on fruit centers to account for the first source of double count. The second part uses these fruit centers and their associations across frames as feature matches in a semantic SfM reconstruction to estimate 3D landmark positions as well as camera poses of each image frame. The third part projects those 3D landmarks back to the image plane of every image in order identify double tracks and address the second source of double count. The fourth part estimates the 3D locations of tree centroids, which are approximated by tree trunks in our implementation. These centroids are used as depth thresholds so only fruits that are closer to the camera than the tree centroids are counted, thus accounting for the third source of double count. Finally, the last part further refines the fruit association across frames, and estimates the final map of fruit landmarks.

\subsection{Tracking in the Image Plane}
Similar to our previous work~\cite{liu2018robust}, we use a combination of the Kanade-Lucas-Tomasi (KLT) optical flow estimator, Kalman Filter, and Hungarian Assignment algorithm to track fruits across image frames. However, we improve upon previous work by defining a different filtering step which fuses both the Faster R-CNN detections and KLT estimates as measurements. 

Each detection from the Faster R-CNN is associated with a center and bounding box. Let $c_{i,k} = [c^{(u)}_{i,k}, c^{(v)}_{i,k}]^\mathrm{T}$ represent the row and column of the center of fruit $i$ in the image coordinate space of image $\mathcal{I}_{k}$, and let $a_{i,k}$ denote the area of the bounding box. We use the KLT tracker to estimate the optical flow $d_{i,k} = [d^{(u)}_{i,k}, d^{(v)}_{i,k}]^\mathrm{T}$ for the fruit at $c_{i,k}$ in order to get the predicted location $\hat{c}_{i,k+1} = c_{i,k} + d_{i,k}$ in image $\mathcal{I}_{k+1}$. After this optical flow prediction step, we denote the overlap proportion of the predicted bounding box of fruit $i$ at predicted position $\hat{c}_{i,k+1}$ with the bounding box of detected fruit $j$ in $\mathcal{I}_{k+1}$ as $\gamma_{k+1}(i,j)$. 

Our tracking task is a \textit{Multiple Hypothesis Tracking} (MHT) problem~\cite{reid1979algorithm}. Given two sets of detections in consecutive images $\mathcal{I}_{k}$ and $\mathcal{I}_{k+1}$, we want to find the cost minimizing assignment which represents the tracks between $\mathcal{I}_{k}$ and $\mathcal{I}_{k+1}$ using  the Hungarian Algorithm~\cite{munkres1957algorithms}. 
Each possible assignment between a detection $i$ in $\mathcal{I}_{k}$ and $j$ in $\mathcal{I}_{k+1}$ is associated with the following cost:
\begin{equation*}
C(i,j,k) = \frac{||\hat{c}_{i,k+1} - c_{j,k+1}||_{2}^{2}}{a_{i,k} + a_{j,k+1}} + \prl{1 - \gamma_{k+1}(i,j)},
\end{equation*}


Once detection $i$ in $\mathcal{I}_{k}$ has been assigned to detection $j$ in $\mathcal{I}_{k+1}$, we have two measurements for its position in image $\mathcal{I}_{k+1}$ from the KLT tracker ($\hat{c}_{i,k+1}$) and the Faster R-CNN detection ($c_{j,k+1}$). We use a Kalman Filter~\cite{kalman1960new} to fuse these two measurements to obtain the final estimates of fruits' positions, which we denote as $p_{i,k}$. Note that $c_{i,k}$ represents the centers of Faster R-CNN detected bounding boxes, while $p_{i,k}$ represents the filtered estimates of the fruit positions.



For every new fruit, we initialize its own Kalman Filter upon first detection. Define the expanded $4 \times 1$ state vector $x_{i,k}$ as:
\begin{equation*}
x_{i,k} = \begin{bmatrix}
           p_{i,k},
           \dot{p}_{i,k}
         \end{bmatrix} ^\mathrm{T}
         = \begin{bmatrix}
           u_{i,k},
           v_{i,k},
           \dot{u}_{i,k},
           \dot{v}_{i,k}
         \end{bmatrix}^\mathrm{T},
\end{equation*}
where we now include $\dot{u}_{i,k}$ the pixel row velocity and $\dot{v}_{i,k}$ the pixel column velocity, both of which have the unit of ($\frac{pixel} {\Delta t_{f}}$), where $\Delta t_{f}$ is the constant time interval between every two frames. Let $\bar{d}_{k} =
\begin{bmatrix}
    \frac{1}{m}\sum_{l=1}^{m}d^{(u)}_{l,k},
    \frac{1}{m}\sum_{l=1}^{m}d^{(v)}_{l,k} 
\end{bmatrix}^\mathrm{T}$ be the average optical flow of all fruits in $\mathcal{I}_{k}$. The initial value for the fruit's position is the center of its Faster R-CNN detected bounding box, and the average optical flow of all fruits in the previous image $\mathcal{I}_{k-1}$:

\begin{equation*}
{{x}}_{i,k} 
    \xrightarrow[\text{}]{\text{initialize}} \begin{bmatrix}
    c_{i,k}\\
	\bar{d}_{k-1}
\end{bmatrix}.\\
\end{equation*}
In using the average optical flow to initialize the fruit's velocity, we exploit the fact that the perceived movement of the fruit in 2D is due to the motion of the camera.

We use the following discrete-time time-invariant linear system model
\begin{equation}
\begin{aligned}
x_{i,k+1} &= \mathbf{A}x_{i,k}  + \omega \nonumber
\end{aligned}
\label{eqn:process_model}
\end{equation}

\begin{equation*}
z_{i,k+1} = \begin{bmatrix}
    \hat{c}_{i,k+1} \\
    c_{j,k+1} \\
    \frac{\bar{d}_{k+1}}{||\bar{d}_{k+1}||} \cdot ||d_{i,k}|| \\
\end{bmatrix} = \mathbf{H} x_{i,k+1} + n
\end{equation*}
where $z_{i,k+1}$ is our $6 \times 1$ measurement vector consisting of the optical flow and Faster R-CNN measurements (assuming that detection $i$ has been associated with detection $j$ in the previous step), as well as an additional velocity measurement that multiplies the magnitude of the fruit's optical flow in the previous image $d_{i,k}$, which approximates the fruit's depth w.r.t. camera, with current normalized optical flow direction $\frac{\bar{d}_{k+1}}{||\bar{d}_{k+1}||}$.


$\mathbf{A}$ is the state transition matrix, $\mathbf{H}$ the observation matrix, $\omega$ the process noise, and $n$ the measurement noise. Both $\omega$ and $n$ are random variables assumed to be drawn from a Gaussian zero-mean distribution. Specifically, in our implementation, those quantities are defined as follows:
\begin{gather*}
\mathbf{A} = \begin{bmatrix}
    \mathbf{I}_{2} & \mathbf{I}_{2} \\
    \mathbf{0} & \mathbf{I}_{2}  \\
\end{bmatrix}, \hspace{0.1in}
\mathbf{H} = \begin{bmatrix}
    \mathbf{I}_{2} & \mathbf{0} \\
    \mathbf{I}_{2} & \mathbf{0} \\
    \mathbf{0}  & \mathbf{I}_{2} \\
\end{bmatrix}, \\
\omega \sim \mathcal{N}(0, \mathbf{Q}),\hspace{0.1in} n \sim \mathcal{N}(0, \mathbf{R})
\end{gather*}
where, $\mathbf{I}_{2}$ is the identity matrix of size $2$, $\mathbf{Q} = diag(6,2,3,1)$ is $4 \times 4$ covariance matrix, and $\mathbf{R} = diag(3,1,1,0.5,1,0.5)$ is $6 \times 6$ covariance matrix. We chose the relative magnitudes for these covariance matrices since the Faster R-CNN measurements are relatively precise while optical flow measurements are sometimes noisy.

Thus, given a state $x_{i,k}$, using the process model in Eqn~\eqref{eqn:process_model}, we will get an \textit{a priori} estimate $\hat{x}^{-}_{i,k+1}$ for image $\mathcal{I}_{k+1}$ given knowledge of the process prior to step $k+1$. Using the measurement $z_{i,k+1}$, we perform the standard Kalman Filter prediction and update steps detailed in~\cite{kalman1960new} to compute the \textit{a posteriori} state estimate $\hat{x}_{i,k+1}$. In this way, we keep propagating the state vector and covariance matrix of every fruit until we lose track of it.

Using the above tracking process, we extract the full tracking history of every fruit to construct a set $\mathbf{M}^{F}$ of the fruit feature matches. If a fruit detection $i$ has been tracked from $\mathcal{I}_{k-n}$ to $\mathcal{I}_{k}$, we add the entire sequence of tracked positions to $\mathbf{M}^F$: 
\begin{equation*}
\mathbf{M}^F = \mathbf{M}^F \cup  \begin{bmatrix}
p_{i,k-n},
p_{i,k-(n-1)},
\hdots,
p_{i,k}
\end{bmatrix} ^\mathrm{T} \\
\end{equation*}

By constructing this set $\mathbf{M}^{F}$, we account for the first kind of double counts which results from detecting the same fruit in consecutive frames.

\subsection{Semantic SfM: Estimate the Camera Poses and the Fruit Landmark Positions from Fruit Feature Matches}\label{SfM}

The most common SfM implementation associates a descriptor with each geometric feature point, and matches them using nearest neighbors in the descriptor space. This descriptor matching process is computationally expensive~\cite{karami2017image}. Also, extra effort is required to localize objects of interest in 3D. 

We propose a semantic SfM which directly uses the frame to frame fruit feature matches $\mathbf{M}^F$ output by the 2D tracking process. As a result, the SfM computation is greatly sped up and meanwhile fruits are directly localized in 3D. The motivation of our fruit-as-feature semantic SfM includes that (1) the fruits are consistently tracked across multiple images and between every pair of images there are sufficient (generally 50 to 100) fruit tracks, and that (2) the frame-to-frame fruit tracking and matching process considers not only the optical-flow-based estimates, but also other properties associated with every fruit including its Fater R-CNN detection, making it more robust to factors such as illumination shifts than just tracking points. We use the COLMAP package \cite{schoenberger2016sfm} \cite{schoenberger2016mvs} as our SfM implementation. The outputs of this SfM step are a set of 3D landmarks $\{\mathcal{L}_i\}$ corresponding to the fruits and a set of camera poses $\{\mathbf{P}_{k}\}$ for each frame. Each landmark $\mathcal{L}_{i}$ has an associated 3D position $\mathbf{X}_{i}$.

The first step is to identify a good initial pair of images to start the SfM reconstruction process. We input our fruit correspondences $\mathbf{M}^F$ as raw feature matches, and then conduct a geometric verification step \cite{schoenberger2016sfm}, which uses Epipolar geometry \cite{hartley2005multiple} and RANSAC \cite{fischler1981random} to determine the best initial pair of images as well their inlier feature matches. These initial images are used to initialize the SfM reconstruction process using two-view geometry  \cite{schoenberger2016sfm} \cite{beder2006determining} \cite{hartley2003multiple}, in which the initial pair of camera poses are estimated and landmarks observed in the initial pair of images are initialized. The Perspective-n-Point (PnP) algorithm \cite{fischler1981random} is then employed to incrementally estimate poses corresponding to the preceding and succeeding images, while multi-view triangulation is conducted to initialize new landmarks \cite{schoenberger2016sfm}.

\begin{figure}[t!]
\vspace*{0.01in}
\centering
\includegraphics[width=3.4in]{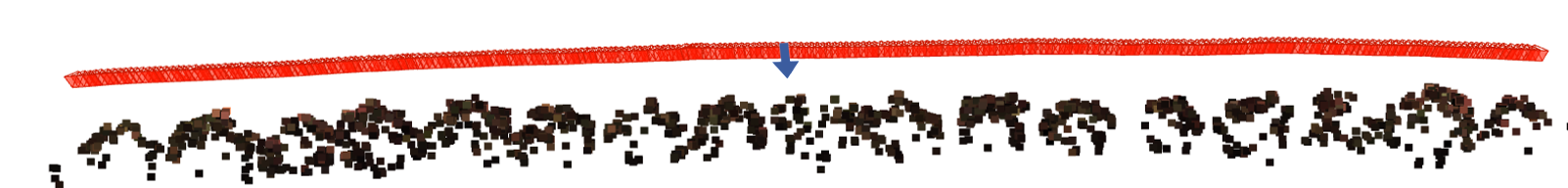}
\caption{\textit{Semantic SfM estimated fruit map and camera poses using one image sequence}. The blue arrow denotes the front direction of the camera. Every point represents a fruit center and every red tetrahedron represents a camera pose. There are 17 trees in this image sequence.}
\label{fig:sfm}
\vspace*{-0.3in}
\end{figure}

While the above process generates initial estimates of camera poses and landmark positions, uncertainties in poses and landmark positions can increase overtime and cause the system to drift. Therefore, an optimization process is needed to correct for these errors. A common approach is to minimize the reprojection error of the landmarks. The reprojection error is defined as
\begin{equation*}
e_{i,k} = ||\hat{p}_{i,k} - p_{i,k}||_2^2,
\end{equation*}
where $\hat{p}_{i,k}$ is landmark $\mathcal{L}_i$'s projection in $\mathcal{I}_{k}$ and $p_{i,k}$ is the actual position of the fruit determined by our Kalman Filter. The projection $\hat{p}_{i,k}$ of landmark $\mathcal{L}_i$ in image $\mathcal{I}_{k}$ with estimated camera pose $\mathbf{P}_{k}$ can be calculated as:
\begin{equation}
\begin{aligned}
\begin{bmatrix}
\hat{u}_{i,k} \\
\hat{v}_{i,k} 
\\
1
\end{bmatrix} &= \mathbf{K} [\mathbf{R}_{k} | \mathbf{T}_{k}] \begin{bmatrix}
\mathbf{X}_i \\
1
\end{bmatrix} \\
\hat{p}_{i,k} &= \begin{bmatrix}
\hat{u}_{i,k},
\hat{v}_{i,k} 
\end{bmatrix}
^\mathrm{T},
\end{aligned}
\label{eq:projection}
\end{equation}
where $\mathbf{K}$ is the $3 \times 3$ camera intrinsic matrix, $\mathbf{R}_{k}$ is the $3 \times 3$ rotation matrix and $\mathbf{T}_{k}$ is the $3 \times 1$ translation vector that defines the camera rotation and translation in the world frame. $\mathbf{R}_{k}$ and $\mathbf{T}_{k}$ can be derived from camera pose $\mathbf{P}_{k}$.

Bundle Adjustment (BA)~\cite{triggs1999bundle} solves the following nonlinear optimization problem:
\begin{equation*}
\min_{\mathbf{P}_k, \, \mathbf{X}_i} \sum_{i=1}^{N} \; \sum_{k=1}^{K} \; \omega_{i,k} \, e_{i,k},
\end{equation*}
where $N$ is the total number of landmarks, $K$ is the total number of images, and $\omega_{i,k}$ is the binary valued variable denoting observability of $\mathcal{L}_i$ in $\mathcal{I}_k$. The minimizing $\{\mathbf{P}^{*}_{k}\}, \{\mathbf{X}_{i}^{*}\}$ are the estimated landmark locations and camera poses from this SfM step. An example of our semantic SfM reconstruction is shown in Fig. \ref{fig:sfm}.

\subsection{Avoid Double Counting of Double Tracked Fruits: Re-associate 3D Landmarks with Detections}\label{association}

The second kind of double counting is also called double tracking, which results from a missed detection or the fruit being occluded in an intermediate frame. As a result, the 2D tracking and Semantic SfM step will potentially generate multiple landmarks for this fruit. A direct way to account for this problem is to compare the distance between landmarks and reject those that coincide. Unfortunately, this approach does not work well, as the fruits are clustered and the SfM reconstruction only provides relative scale. It is difficult to choose an absolute threshold that determines whether two landmarks coincide. 

Instead, we approach this problem by re-associating every 3D landmark with 2D Faster R-CNN detections, and discarding landmarks which are not associated to any detection. We sequentially project the landmarks back to every image to obtain $\hat{p}_{i,k}^{\mathcal{L}}$ using Eqn.~\eqref{eq:projection}, and match these projections with the detections using a second Hungarian assignment. The cost function of this Hungarian assignment is designed to account for the age of the landmark, or the number of previous images where it has been observed. Using this cost function, for two landmarks corresponding to the same fruit, the older landmark will have lower cost and be matched with the fruit detection, and the newer landmark will be discarded, thus avoiding double counting.
%
%

In addition to the 3D position $\mathbf{X}_{i}$, we associate four additional attributes to landmark $\mathcal{L}_{i}$. The first attribute is a bounding box with area $a_{i}^{\mathcal{L}}$ corresponding to the Faster R-CNN bounding box in the last frame where landmark $\mathcal{L}_i$ is observed. The second attribute is an observability history $(\omega_{i,0}, \dots \omega_{i,K})$ which records the landmark $\mathcal{L}_i$'s observability in every frame. The third attribute is a depth $\lambda_{i,k}$ representing the depth of $\mathcal{L}_i$ w.r.t. the camera center of $\mathcal{I}_{k}$, i.e. the z-axis value of the landmark in the camera coordinate frame. The fourth attribute is the age $O_i$ defined by the number of images where $\mathcal{L}_{i}$ has been observed up until the current image.

%
%
Using these attributes, we define the following Hungarian algorithm cost function for associating landmark $\mathcal{L}_{i}$ with Faster R-CNN detection $j$ in image $k$:

\begin{equation*}
C^{\mathcal{L}}(i,j,k) = \frac{||\hat{p}^{\mathcal{L}}_{i,k} - c_{j,k}||_{2}^{2}}{a^{\mathcal{L}}_{i}  + a_{j,k}} + \prl{1 - \gamma_{k}(i,j)} + C^{\mathcal{L}}_O(s,t).
\end{equation*}
$C^{\mathcal{L}}_O(s,t)$ is an age cost defined as
\begin{equation*}
C^{\mathcal{L}}_O(s,t) = w_{0} \times max \prl{0 ,  \prl{1 - ({{O}_{i} - O_0}) / {O_0}}}.
\end{equation*}
$O_0$ is the threshold for age, which we chose to be 7. $w_O$ is the age cost weight, chosen as 0.5, which controls the contribution of the age cost to the total cost. Conducting this global data association is relatively cheap in computation since our fruits are sparse.

\begin{figure}[t!]
\vspace*{0.05in}
\centering
\includegraphics[width=2.0in]{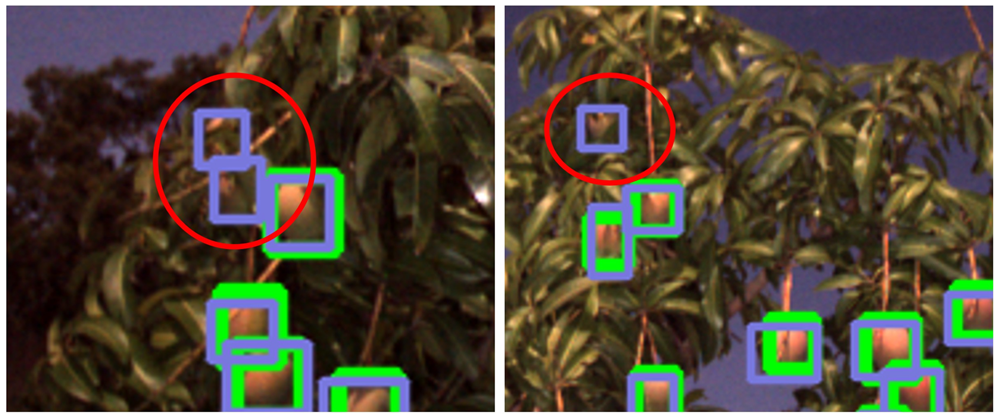}
\caption{{\textit{Projections of fruit landmarks (purple boxes) and Faster R-CNN fruit detections (green boxes)}}. The fruit-landmark-based approach can also improve the fruit detection results, by estimating the positions of highly occluded fruits (in red circles).}
\label{fig:improve_detection}
\vspace*{-0.25in}
\end{figure}

Besides avoiding double counting and improving data association, the landmark projections also help to improve fruit detection in 2D images, as shown in Fig.~\ref{fig:improve_detection}

\subsection{Avoid Double Counting from Two Opposite Tree Sides: Compare the Fruit Landmark with the Tree Centroid}

The third kind of double counting is caused by using two separate image sequences recorded from two opposite sides of a tree row to count fruits. If these two views of a single row are captured consecutively, the SfM reconstruction algorithm should be able to combine both views into a single point cloud by estimating the pose of the camera as it turns around and faces the other direction. However, our dataset first captures all rows facing one direction (east), and then turns around to capture those rows facing the other direction (west). As a result, the SfM reconstruction process generates two separate point clouds for each row, and we need to integrate them together in order to prevent double counting fruits that are visible from both sides.

\begin{figure}[t!]
\centering
\includegraphics[height=0.9in]{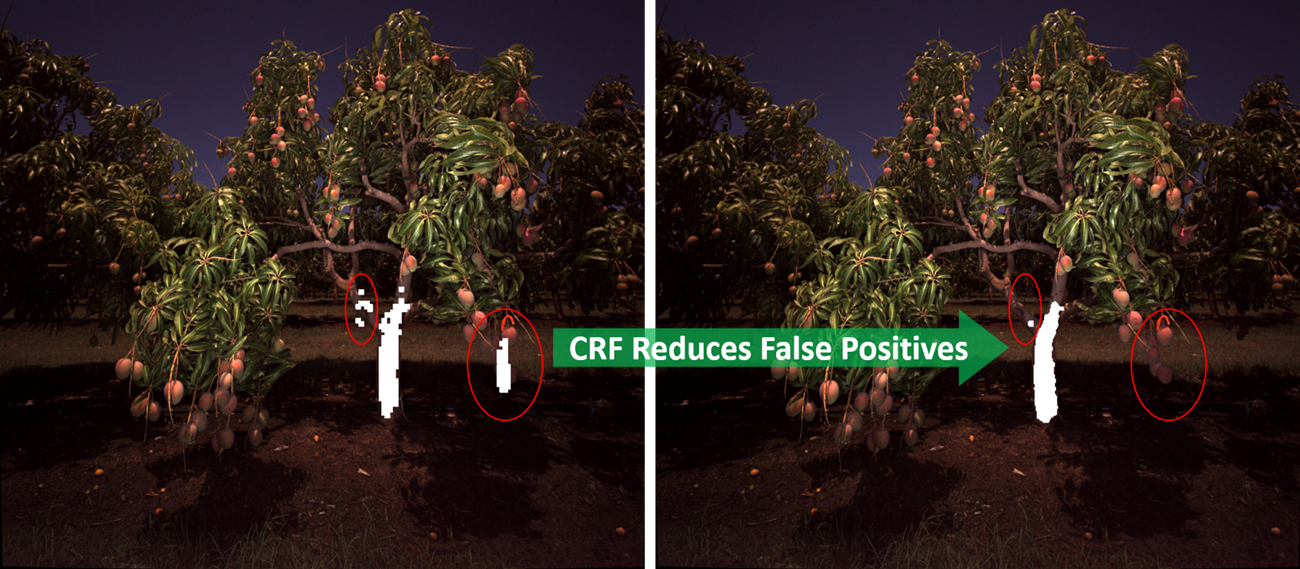}
\vspace*{-0.05in}
\caption{{\textit{Comparison of tree trunk segmentation results before (left) and after (right) adding dense CRF}}. White masks are predicted trunk regions. Adding dense CRF reduces the false positives significantly, which is important for accurately estimating the trunk position.}
\label{fig:trunk_seg}
\vspace*{-0.25in}
\end{figure}

We approach this double counting problem by using the tree centroid, which is represented by the tree trunk, to separate the tree into two parts. We then only count the fruits that lie on the closer side of the tree trunk. In order to estimate the location of the trunk, we track Shi-Tomasi corners on the trunks~\cite{shi1993good}, which is a modified version of Harris corners~\cite{harris1988combined}. We need the corner features to lie on the trunk, and therefore we use the context extraction network, based on the Fully Convolutional Network (FCN) structure \cite{long_shelhamer_fcn} to segment trunks in every image. The context extraction network takes in an image of size $h' \times w' \times 3$ and outputs a score tensor of size $h' \times w' \times n_c$ where $n_c$ is the number of object classes. A dense CRF~\cite{Krahenbuhl2012} is added to refine the network output, which forces consistency in segmentation and sharpens predicted edge, as shown in Fig.~\ref{fig:trunk_seg}. This is especially important for the next tracking stage because we want all extracted corners to be on the trunk.


For a trunk $l$, we first manually choose a start frame $\mathcal{I}_{s}$ according to the segmentation network's segmentation results. We only keep segmented trunk masks which lie in the middle $\frac{1}{3}$ of the image, ignoring the left $\frac{1}{3}$ and right $\frac{1}{3}$. This choice is because that the illumination of this part of image is most sufficient, and that there is only one trunk in the middle of every image, thus avoiding tracking non-target trunks. We extract $m$ corner points $\left \{ {{T_{1,s}}, {T_{2,s}}, {T_{m,s}}} \right \}$ inside the trunk region, and use the KLT tracker to track them across multiple frames from $\mathcal{I}_s$  to $\mathcal{I}_{s+b}$. For every tracked corner point, we obtain a set of point correspondences from $\mathcal{I}_s$ to $\mathcal{I}_{s+b}$, and add it to the trunk feature match set.

\begin{figure}[t!]
\centering
\includegraphics[height=0.9in]{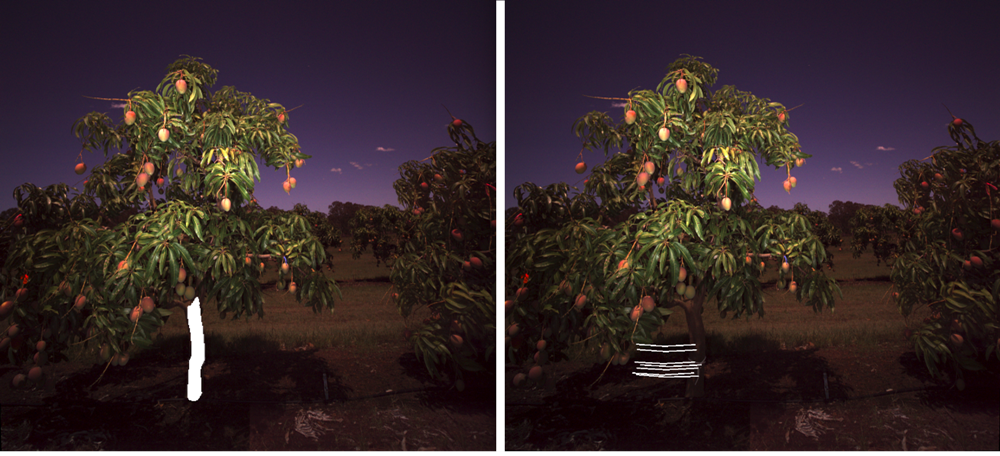}
\vspace*{-0.05in}
\caption{{\textit{Trunk segmentation (left) and tracking (right)}}. Since trunks are much larger than fruits, instead of directly tracking the whole trunk, we extract corners within the predicted trunk mask and track them. The white lines are the trajectories of extracted corner points.}
\label{fig:trunk_track}
\vspace*{-0.25in}
\end{figure}

We set b to be 3, i.e., we track every point across $4$ frames, which achieves a good trade-off between robustness in tracking and accuracy in triangulation. Fig.~\ref{fig:trunk_track} shows the tracking process.

Using the pose estimates from $\mathcal{I}_s$ to $\mathcal{I}_{s+b}$, we conduct a multi-view triangulation for those corner points and calculate their depth w.r.t. the camera center of every frame. Considering that most false positive pixels for trunk segmentation lie on closer objects such as leaves or fruits (because they occlude the trunks), we represent the depth $\lambda ^ T_{l,s}$ of the trunk $l$ at $\mathcal{I}_s$ using the third quartile of the depth of all corner points at $\mathcal{I}_s$, i.e., $\left\{ {\lambda_{T_{1,s}}, \lambda_{T_{2,s}}, ..., \lambda_{T_{m',s}}} \right\}$. 
%
%
For every fruit landmark, before counting it in $\mathcal{I}_k$, we look back 15 frames, and calculate the depth of trunk $\left\{{\lambda ^ T_{l,k-15}, \lambda ^ T_{l,k-14} ... \lambda ^ T_{l,k} }\right\}$ and the depth of the landmark $\mathcal{L}_i$, $\left\{{\lambda _ {i,k-14}, \lambda  _ {i,k-15} ... \lambda  _ {i,k}}\right\}$. We then employ the following voting system to decide whether the fruit landmark is before the tree centroid: for $f$ in $\left\{{k-15, k-14, ..., k} \right\}$, if $\lambda _ {i,f} < \lambda ^ T_{l,f}$, we add a before-centroid vote; otherwise, we add an after-centroid vote. We will count this landmark only if its before-centroid votes are more than its after-centroid votes.

\subsection{Refine Final Data Association and Final SfM}
The final step of our algorithm is to estimate the final map of fruits using the semantic SfM introduced in section \ref{SfM}, with the fruit feature matches after re-association in section \ref{association}. To guarantee the quality of reconstruction, the position difference of the matched fruit detections in two consecutive images is regarded as an initial guess of this fruit's displacement between the two images, and the KLT algorithm is used to find the optimal displacement based on this initial guess. For every fruit, we conduct this refinement consistently through all frames where it has been tracked. The position of the fruit $i$ at the first tracked frame $\mathcal{I}_{k-n}$ is defined as its Faster R-CNN detected bounding box center $p^*_{i,k-n} = c_{i,k-n}$. Then the refined optical flow of the fruit $i$ in the image $\mathcal{I}_{k-n}$ is calculated as $d^*_{i,k-n} = [d^{*(u)}_{i,k-n}, d^{*(v)}_{i,k-n}]^\mathrm{T}$. The refined position of fruit $i$ in the image $\mathcal{I}_{k-(n-1)}$ is defined as $p^*_{i,k-(n-1)} = p^*_{i,k-n} + d^*_{i,k-n}$. This is computed iteratively until we lost track of this fruit at $\mathcal{I}_{k}$. Therefore, the refined data association built by this fruit is added to the refined fruit feature matches set: 
\begin{equation*}
{\mathbf{M}_k^{F}}^* = {\mathbf{M}_k^{F}}^* \cup  \begin{bmatrix}
p^*_{i,k-n},
p^*_{i,k-(n-1)},
...,
p^*_{i,k}
\end{bmatrix} ^\mathrm{T}
\end{equation*}

With this refined fruit feature match set ${\mathbf{M}_k^{F}}^*$, our final SfM output is shown in Fig.~\ref{fig:sfm}.

\section{Results and Analysis}


\begin{figure*}[t!]
\centering
\includegraphics[width=6.5in]{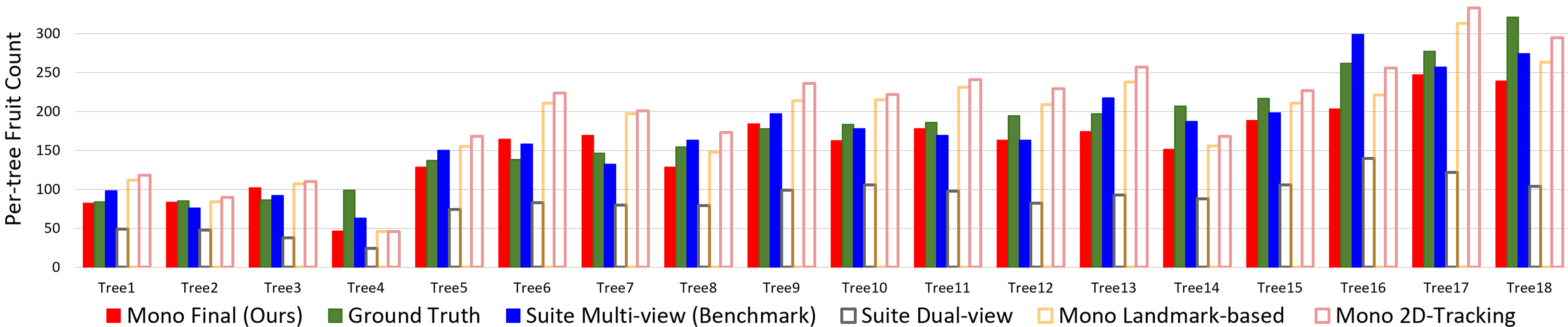}
\vspace*{-0.05in}
\hspace*{0.05in}
\caption{\textit{Comparison of per-tree count results from all algorithms and field (ground truth) count}. 18 ground-truth trees are sampled from all 10 tree rows in the orchard. The X-axis is the tree index, where trees are ranked from smallest to largest according to their ground-truth counts. The Y-axis is the per-tree fruit count. Both our \textit{Mono Final} algorithm and the benchmark \textit{Suite Multi-view} algorithm well estimate the field counts for most ground truth trees. A significant improvement can be seen from our 2D tracking outputs to landmark-based outputs, and from landmark-based outputs to our final outputs (count outputs from different stages of our pipeline are denoted in Fig. \ref{fig:pipeline}). \textit{Suite Dual-view} algorithm is severely under-counting, mainly because that it only uses two opposite images of every tree to get the count, indicating the advantages of counting from image sequences.}

\label{fig:all_result_comp_bar}
\vspace*{-0.15in}
\end{figure*}

\begin{figure*}[t!]
\centering
\includegraphics[width=6.5in]{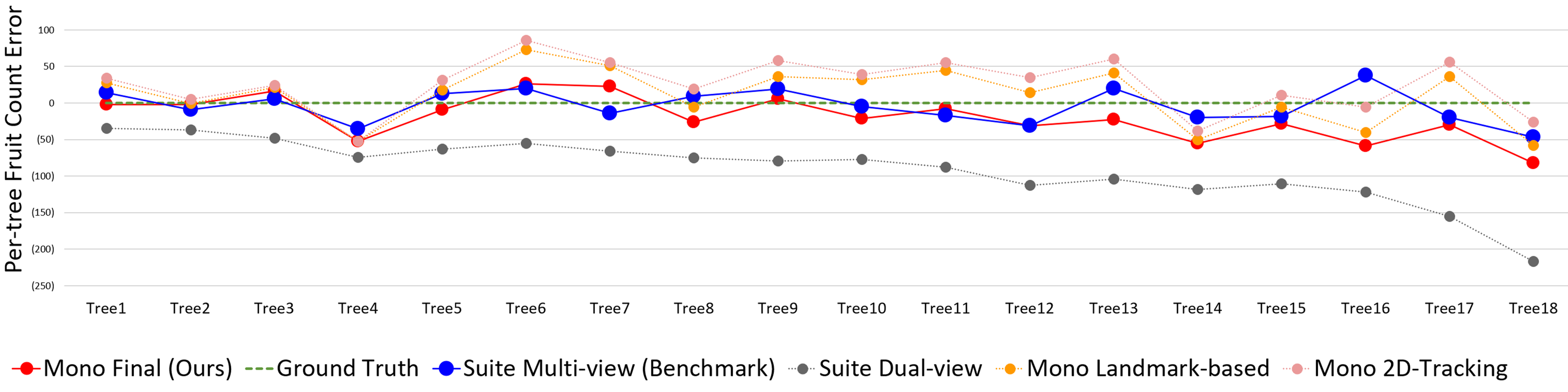}
\vspace*{-0.10in}
\caption{\textit{Comparison of per-tree count error against the field count}. The X-axis is the tree index. The Y-axis is the per-tree fruit count error against the field count (parentheses indicate negative). On average every tree has 175 mangoes. Our \textit{Mono Final} algorithm and the benchmark \textit{Suite Multi-view} algorithm have least errors against the field count. Both algorithms are on average slightly under-counting, which is expected since we are comparing against the field count that includes fully occluded fruits. It is notable that all algorithms show a similar trend in their counting errors, which reflects the variance in the occlusion conditions of ground truth trees.}
\label{fig:all_result_comp}
\vspace*{-0.25in}
\end{figure*}


\begin{figure}[t!]
\centering
\includegraphics[width=3.4in]{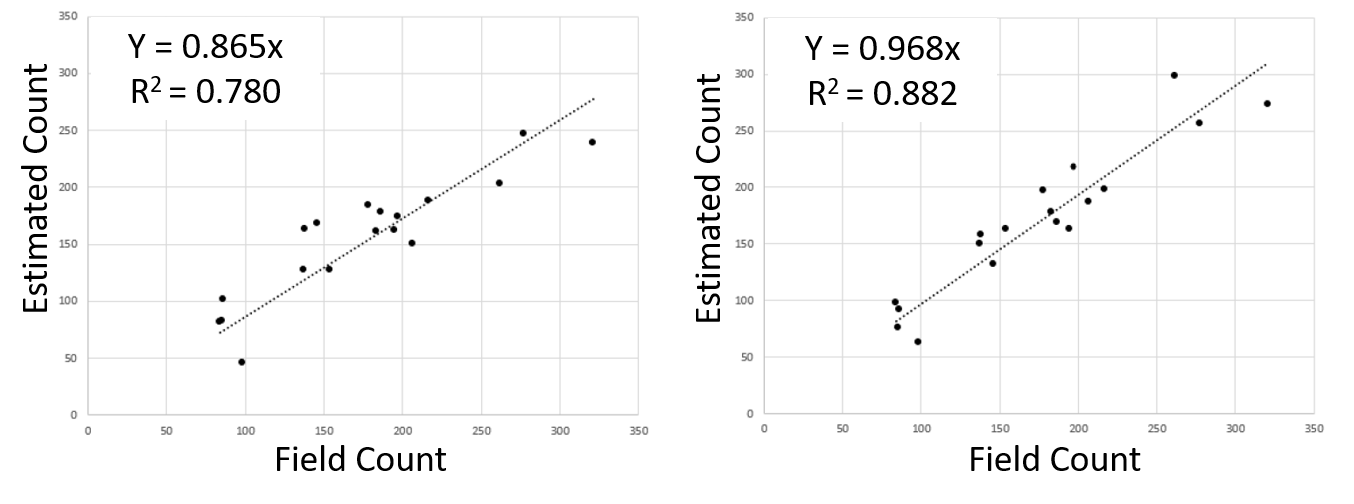}
\vspace*{-0.1in}
\caption{\textit{Comparison of linear regression models for our \textit{Mono Final} algorithm (left) and the benchmark \textit{Suite Multi-view} algorithm (right)}. The X-axis is the field count and the Y-axis is the estimated count. Every dot represents a tree. For both methods, the slope is reasonable and $R^2$ are within a relative high precision range.}
\label{fig:result_scatter}
\vspace*{-0.15in}
\end{figure}

\begin{figure}[t!]
\begin{center} 
 \setlength\extrarowheight{5pt}
 \resizebox{0.4\textwidth}{!}
 {\begin{tabular}{||c | c | c ||} 
 \hline
 Measure &  Mono Final &  Suite Multi-view\\ [0.6ex] 
 \hline\hline
Per-tree Count Error Mean &  \textbf{27.8}  & \textbf{19.8}\\
  \hline\hline
Per-tree Count Error Std Dev & \textbf{29.6} & \textbf{22.9}\\
    \hline\hline
\end{tabular}}
\end{center}
\vspace*{-0.1in}
\caption{\textit{Per-tree count error mean and standard deviation of our Mono Final algorithm and the benchmark Suite Multi-view algorithm}.}
\label{fig:result comp table}
\vspace*{-0.25in}
\end{figure}

%
%
%
%
In this section, we compare both the estimated count output from our monocular camera system and the count output of a benchmark sensor-suite-based system from~\cite{stein2016image} that uses the camera, LiDAR and GPS/INS system against the ground truth field count. The benchmark system uses the same Faster R-CNN as ours to detect fruits in images, the GPS/INS/RTK system to estimate the camera poses, and the 3D LiDAR point cloud to estimate the tree centroids and tree masks. Two algorithms are included in this benchmark system, the Suite Multi-view algorithm and the Suite Dual-view algorithm. The Suite Multi-view algorithm uses all sequence images, while the Suite Dual-view algorithm uses only 2 images from 2 opposing views (one facing east and one facing west). Although our monocular camera system only utilizes the 2D images, the results show that it has comparable performance with the Suite Multi-view algorithm.

Our data set was collected using an unmanned ground vehicle (UGV) built by the Australian Centre for Field Robotics (ACFR) at The University of Sydney. It has a 3D LiDAR and GPS/INS which is capable of real-time-kinematic (RTK) correction. In addition, it has a Prosilica GT3300C camera with a Kowa LM8CX lens, which captures the RGB images of size $3296 \times 2472$ pixels (8.14 megapixels) at 5Hz~\cite{stein2016image}. 

The data set was collected on December 6, 2017, from a single mango orchard at Simpson Farms in Bundaberg, Queensland, Australia. We manually counted 18 trees in the field as ground truth. The 18 ground-truth trees were chosen from all 10 rows of trees in the orchard, to maximise variability of NDVI (and by extension yield) from multi-band satellite data. The trees vary in size and occlusion conditions.

Fig.~\ref{fig:all_result_comp_bar}, Fig.~\ref{fig:all_result_comp} and Fig.~\ref{fig:result comp table} show the counting results of our monocular camera system and the sensor suite system. To measure and compare the per-tree counting performance, we manually mask the target trees in our algorithm.

Fig.~\ref{fig:all_result_comp_bar}, and Fig.~\ref{fig:all_result_comp} also depict the improvement in counting results from the three stages of our pipeline as shown in Fig.~\ref{fig:pipeline}. Our raw 2D-tracking count and raw landmark-based count both have over-counting trends, however, the landmark-based one better estimates the ground truth in most cases. This indicates the improvement introduced by landmark-based double tracks rejection. After using tree centroids to get rid of two-side double counts, our final output is much more robust and accurate. In addition, Suite Dual-view algorithm undercounts, mainly because it only uses two opposite images of every tree to get the count. This indicates that multi-view sequence images cover more fruits, and thus it is a more desirable data collection approach.




Fig.~\ref{fig:result_scatter} shows the linear regression models fitted for our monocular-camera-based algorithm's final output and sensor-suite-based multi-view algorithm's output.  A slope of 1 indicates that the estimated count is proportional to the field count, and a high $R^{2}$ value indicates that the linear model on the field counts is a good fit for estimated counts. For the monocular camera system, the slope is $0.86$ compared to $0.97$ for the sensor suite system, and the $R^{2}$ value is $0.78$ compared to $0.88$. The linear regressions show that most of the data points corresponding to high field counts lie below the unit diagonal line, which corresponds to our previous observation that undercounting occurs due to the highly occluded fruits. The metrics indicate that both systems are performing well.


Most of the difference in performance between the monocular camera system and the sensor suite system comes from trees with higher counts, indicating that the sensor suite multi-view algorithm can better handle occluded fruit. We would expect both algorithms to handle fruit occlusions equally well since they are using the same Faster R-CNN detection network. However, the occluded fruit can cause a performance difference due to the third source of double counting that results from combining point clouds from the two different viewpoints. Due to higher occlusion, from a given side of the tree, it is more likely that a fruit may lie further than the trunk centroid, but is not visible from the other side of the tree. Throwing away this landmark just because it is on the wrong side of the tree would be too aggressive of a strategy, and may be the cause of the larger amount of undercounting. 

One solution to this problem is a more sophisticated algorithm to integrate the two point clouds. As previously mentioned though, if the data was collected so that the two views of the same row were consecutive, our monocular algorithm would not need the trunk centroid rejection step since the SfM reconstruction algorithm would be able to integrate both views into the same point cloud.

One strength of our algorithm is that by using semantic SfM on fruit features rather than SIFT features, we achieve a much faster algorithm. On a 4 core i7 CPU for a 1000 frame video, our SfM reconstruction takes about 5 minutes compared to 10 hours for traditional SIFT based SfM (note that this does not include 2D fruit tracking computation, since for our task fruit tracking is needed whether or not it is used for SfM). This dramatic speed increase is because that, by first estimating the fruit data association across frames, we bypass the computationally expensive feature extraction and matching process based on nearest neighbors on the SIFT descriptors.

\section{Conclusion and Future Work}
We presented a monocular fruit counting pipeline that paves the way for yield estimation using commodity smartphone technology. Such a fruit counting system has applications in a wider variety of farm environments where cost and environment constraints prevent the usage of high cost and larger sensors. Our pipeline begins with detecting fruits and tree trunks using CNNs. These detected fruits and trunks are then tracked across images. A semantic SfM is then used to directly convert 2D tracks of fruits and trunks into 3D landmarks. With these landmarks, the double counting scenarios are identified and the fruit data association is refined. Finally, the total fruit count and map are estimated. We evaluated our monocular system on a mango dataset against the actual field count as well as a sensor suite algorithm that uses a monocular camera, 3D LiDAR and GPS/INS system. 

When designing a fruit counting system, there is a tradeoff between hardware complexity and software complexity. We have identified modes (trees with high fruit counts) where the monocular camera only algorithm underperforms the sensor suite algorithm. Further experimentation is required to understand this tradeoff curve, and identify more possible failure cases to be improved upon. Despite the restriction to low cost sensors, we have demonstrated that our monocular system has comparable performance to the system based on expensive sensor suite, and it is a step towards the ultimate goal of a low cost, robust, lightweight fruit counting system.

\bibliographystyle{IEEEtran}
\bibliography{ref.bib}

\end{document}